\documentclass[10pt,twocolumn,letterpaper]{article}
\usepackage{cvpr}
\usepackage{times}
\usepackage{epsfig}
\usepackage{graphicx}
\usepackage{amsmath}
\usepackage{amssymb}
\usepackage{authblk}
\usepackage{xcolor}
\usepackage{xspace}
\usepackage{multirow}
\usepackage{bbm}
\usepackage{multirow}
\usepackage{subcaption}
\usepackage{microtype}
\usepackage{algorithm}
\usepackage[noend]{algpseudocode}
\usepackage{tabularx}
\usepackage{makecell}
\usepackage{breqn}
\usepackage{booktabs}
\usepackage{url}
\usepackage[colorlinks=true,citecolor=blue,breaklinks=true,bookmarks=false]{hyperref}

\cvprfinalcopy % *** Uncomment this line for the final submission

\def\cvprPaperID{110} % *** Enter the CVPR Paper ID here

\newcommand{\BT}{{\sc BigTime}\xspace}
\newcommand{\BTShort}{{\sc BT}\xspace}

\newcommand{\Images}{\mathcal{I}}
\newcommand{\Reflectances}{\mathcal{R}}
\newcommand{\Shadings}{\mathcal{S}}
\newcommand{\Energy}{\mathcal{E}}
\newcommand{\Mask}{M}
\newcommand{\acronym}{APWLS\xspace}

\newcommand{\Lrc}{\mathcal{L}_{\mathsf{consistency}}}
\newcommand{\Limrec}{\mathcal{L}_{\mathsf{reconstruct}}}
\newcommand{\Lrsm}{\mathcal{L}_{\mathsf{rsmooth}}}
\newcommand{\Lssm}{\mathcal{L}_{\mathsf{ssmooth}}}
\newcommand{\Lssmi}{L_{\mathsf{ssmooth}}}

\newcommand{\wrc}{w_1} % {w_{\mathsf{rc}}}
\newcommand{\wrsm}{w_2} % {w_{\mathsf{rsm}}}
\newcommand{\wssm}{w_3} % {w_{\mathsf{ssm}}}

\newcommand{\median}{\mathrm{median}}
\newcommand{\vmedian}{v^{\mathsf{med}}}
\newcommand{\vmediannorm}{\overline{v^{\mathsf{med}}}}
\newcommand{\vmediannormpq}{\overline{v^{\mathsf{med}}_{pq}}}
\newcommand{\vmedianweight}{\lambda^{\mathsf{med}}}
\newcommand{\vmediannormweight}{\overline{\lambda^{\mathsf{med}}}}

\newcommand{\fp}{\mathbf{f}_p}
\newcommand{\fq}{\mathbf{f}_q}

\newcommand{\norm}[1]{\left\lVert#1\right\rVert}

\newenvironment{packed_enum}{
\begin{enumerate}
  \setlength{\itemsep}{1pt}
  \setlength{\parskip}{2pt}
  \setlength{\parsep}{0pt}
}{\end{enumerate}}

\makeatletter
\def\BState{\State\hskip-\ALG@thistlm}
\makeatother

\begin{document}

%%%%%%%%% TITLE
\title{Learning Intrinsic Image Decomposition from Watching the World}

\author{Zhengqi Li \qquad Noah Snavely \\
Department of Computer Science $\&$ Cornell Tech, Cornell University}

\maketitle
%\thispagestyle{empty}

%%%%%%%%% ABSTRACT
\begin{abstract}
Single-view intrinsic image decomposition is a highly ill-posed
problem, and so a promising approach is to learn from large amounts of
data. However, it is difficult to collect ground truth training data
at scale for intrinsic images. In this paper, we explore a different
approach to learning intrinsic images: observing image sequences over
time depicting the same scene under changing illumination, and
learning single-view decompositions that are consistent with these
changes. This approach allows us to learn without ground truth
decompositions, and to instead exploit information available from
multiple images when training. Our trained model can then be applied
at test time to single views. We describe a new learning framework
based on this idea, including new loss functions that can be
efficiently evaluated over entire sequences. While prior
learning-based
% intrinsic image
methods achieve good performance on
specific benchmarks, we show that our approach generalizes well to
several diverse datasets, including MIT intrinsic images, Intrinsic
Images in the Wild and Shading Annotations in the
Wild.\footnote{Project at: {\scriptsize
    \url{http://www.cs.cornell.edu/projects/bigtime/}}}

\end{abstract}

%%%%%%%%% BODY TEXT
\section{Introduction}

Intrinsic image decomposition is the problem of factorizing an input
image $I$ into a product of a reflectance image and a shading image: $I
= R \cdot S$.
While the vision community has seen significant advances in
single-image intrinsic image decomposition, it remains a challenging,
highly ill-posed problem. Hence, the use of machine learning for this task is an appealing prospect. Unfortunately, it is also difficult to gather direct ground
truth training data. Previous work has collected ground truth via
painting objects~\cite{grosse2009ground}, synthetic
renderings~\cite{Butler:ECCV:2012, chang2015shapenet}, and manual
annotation~\cite{bell2014intrinsic,kovacs2017shading}, but each of
these methods has significant limitations.

\begin{figure}[t]
  \centering
	\begin{tabular}{@{\hspace{0.1em}}c@{\hspace{0.1em}}}
        \includegraphics[width=1.0\columnwidth]{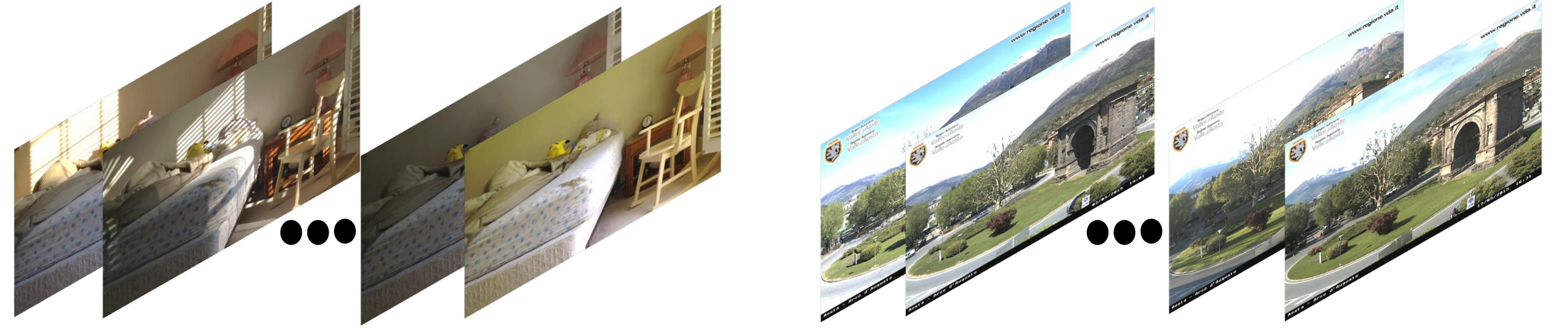}  \vspace{-0.1em} \\ 
        {\small {\bf Training}: we learn from unlabeled indoor and outdoor videos.}  \\ 
        \includegraphics[width=1.0\columnwidth]{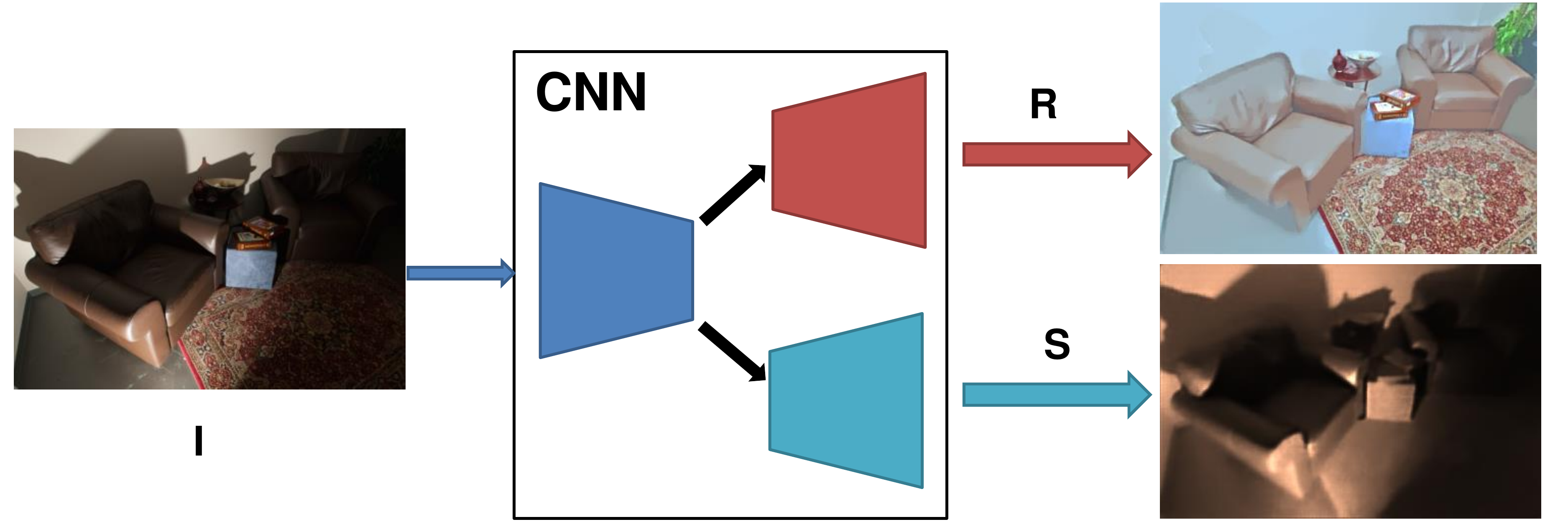} \vspace{-0.1em} \\
        {\small {\bf Testing}: our CNN produces intrinsic images from a single photo.} \vspace{-0.5em}
    \end{tabular} 
	   \caption{
             \label{fig:teaser} To train, our method learns from unlabeled
            videos with fixed viewpoint but varying illumination
            (top).  At test time (bottom), our network produces an
            intrinsic image decomposition $(R, S)$ from a single image
            $I$.} \vspace{-0.5em}
\end{figure}

Inspired by how humans can learn by simply observing the world and
formulating consistent explanations, we consider an alternative,
readily available source of training data for learning intrinsic
images: image sequences from the Internet for which the viewpoint is
fixed but illumination varies. Based on this idea, we introduce \BT
(\BTShort), a large dataset of time-lapse image sequences.
While the sequences in \BTShort do not provide ground truth, they
allow us to incorporate useful constraints during training, by
specifying that the model should predict outputs {\em consistent with
  the sequence}.
While we train on image sequences, our model can apply to a {\em
  single} image at inference time, as illustrated in
Figure~\ref{fig:teaser}.
%%%%%%%

Although a number of prior methods estimate intrinsic images from
sequences, our concept is quite different: we train on sequences, but learn to
infer decompositions from single views. In a sense, our method lies
between optimization-based intrinsic images methods and machine
learning approaches.  In particular, our training loss incorporates
priors similar to those of optimization-based approaches, but in a
feed-forward prediction framework.

To fully utilize the information present in image sequences, we also
introduce two new methods for computing losses over whole sequences,
and show how to efficiently implement these losses inside a deep
network.  The first is an {\em all-pairs weighted least squares} loss
that considers all pairs of images.  The second is a {\em dense,
  spatio-temporal smoothness} loss that jointly considers all of the
pixels in the entire sequence. While we use these losses for training
intrinsic images, they could also be applied to other problems that
involve image sequences, such as video segmentation.

In our evaluation, our method yields competitive or superior
performance on two standard real-world benchmarks, IIW and SAW, even
when trained on \BTShort \emph{without} access to annotations from
those datasets. We further show improved results on the MIT intrinsic
images dataset, even compared to learning methods that utilize full
supervised ground truth.

\section{Related work}\label{sec:related}

\noindent{\bf Intrinsic images through optimization.}  Intrinsic
images has been studied for nearly fifty years,
often within an optimization framework.
Because the problem is ill-posed, additional priors must be applied.
For instance, the seminal Retinex algorithm~\cite{land1971lightness}
assumes large image gradients correspond to changes in reflectance,
while smaller gradients are due to shading.
Subsequently, many different priors have been proposed to guide the
decomposition~\cite{shen2008intrinsic, zhao2012closed,
  rother2011recovering,shen2011intrinsic,garces2012intrinsic}, and
many new optimization tools, such as inference in dense CRFs, have
been deployed~\cite{bell2014intrinsic}.
Some recent approaches make use of surface normals from RGB-D
cameras~\cite{chen2013simple,barron2013intrinsic,jeon2014intrinsic}.
Surface normals can improve shading estimates, but such methods assume
depth maps are available during optimization.

\smallskip
\noindent{\bf Intrinsic images from multiple observations.}
A number of methods, starting with Weiss~\cite{weiss2001deriving},
estimate intrinsic images from time-lapse sequences by
assuming constant reflectance but varying shading over time~\cite{matsushita2004estimating,sunkavalli2007factored,hauagge2013photometric,laffont2012coherent,laffont2015intrinsic}.
Such an approach is similar to our training regime, although a crucial
distinction is that once our model is trained, we can run it on a
single image. These methods rely on priors derived from statistics of
image sequences or lighting sources.
We found that in practice these methods require $a)$ a large number of
input images and $b)$ images taken in outdoor or
controlled laboratory environments. In contrast, our method can learn
from much shorter and less controlled sequences.

\smallskip
\noindent {\bf Intrinsic images via supervised learning.}  Barron and
Malik~\cite{barron2015shape} proposed a unified learning-based method
that incorporates a number of complex priors on shape, albedo, and
illumination. However, their method only applies to single objects and
does not generalize well to real-world scenes. Recently, several
approaches use deep learning to predict albedo and shading via direct
supervision. These methods train on the synthetic
Sintel~\cite{kim2016unified, butler2012naturalistic}, object-centric
MIT~\cite{grosse2009ground} or synthetic ShapeNet
datasets~\cite{chang2015shapenet, janner2017intrinsic}.
However, Sintel and ShapeNet are highly synthetic datasets, and
networks trained on them do not generalize well to real-world
scenes. The MIT dataset consists of real images, but these images
depict objects captured in the lab, not realistic scenes, and the
dataset contains just 20 objects with ground truth.

Recently, two datasets have been created for real-world
scenes. Intrinsic Images in the Wild (IIW)~\cite{bell2014intrinsic} is
a dataset of sparse, human-labeled relative reflectance
judgments. Shading Annotations in the Wild
(SAW)~\cite{kovacs2017shading} similarly contains sparse shading annotations.Several methods
\cite{zhou2015learning,zoran2015learning,narihira2015learning,kovacs2017shading}
train CNNs on sparse annotations from IIW/SAW and use the predictions
as priors for intrinsic images. However, it is difficult to collect
such annotations at scale, especially for shading relationships, which
can be challenging to perceive.  Further, these datasets are limited
to sparse annotations. We propose an alternative form of training data
that is much easier to capture and provides full-image constraints.

\section{Overview and network architecture}\label{sec:overview}

\begin{figure}[t]
 \centering 
 \includegraphics[width=1.0\columnwidth]{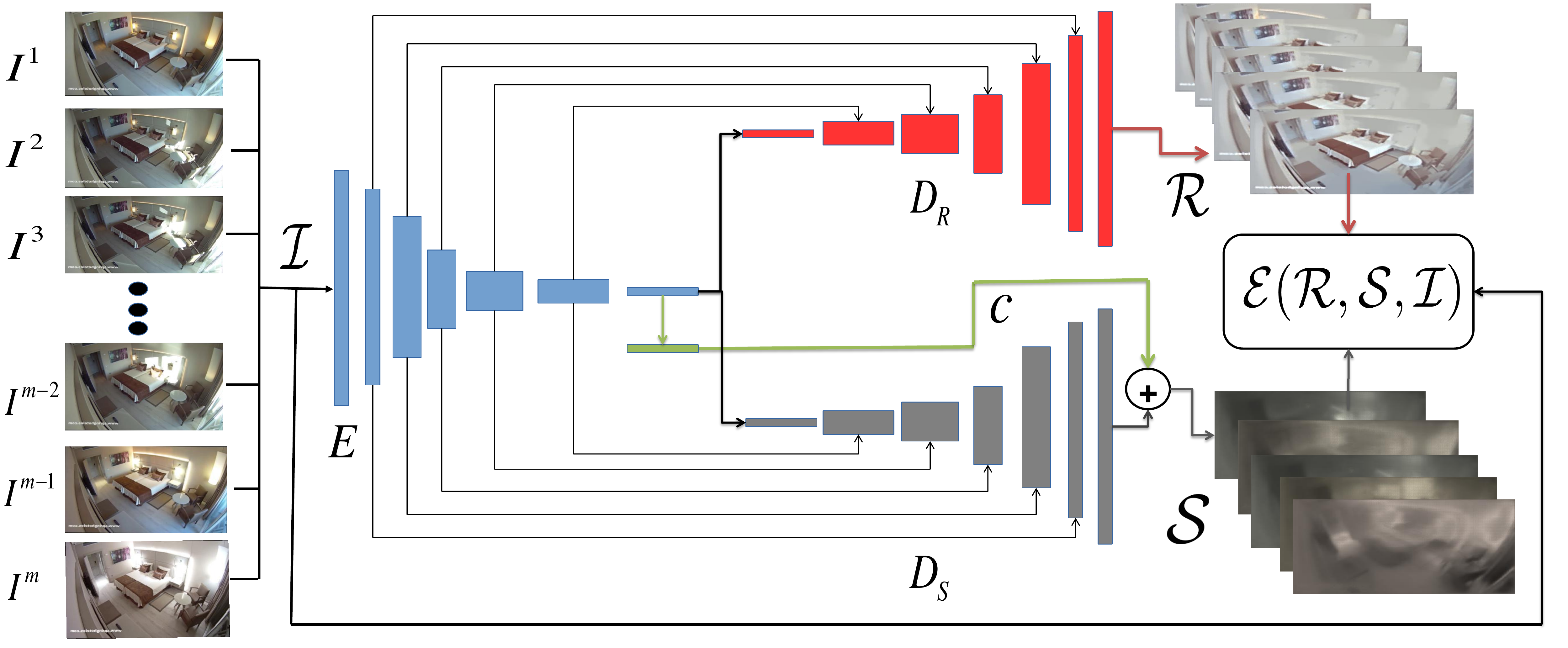} \vspace{-0.5em}
 \caption{\textbf{System overview and network.}  During training, our
   network input is an image sequence $\Images$, and the outputs are
   reflectance images $\Reflectances$ and shading images $\Shadings$
   for the sequence. Each block in the network depicts a
   convolutional/deconvolutional layer. $E$ is an encoder, and $D_R$
   and $D_S$ are decoders for the reflectance and shading images. For
   the innermost feature maps, we have one side output $c$
   representing the illumination color. $\mathcal{E}$ is an energy
   function measuring the cost of the
   decomposition.\label{fig:network} } \vspace{-0.5em}
\end{figure}

Our work makes two main contributions: a new dataset, \BT, of image
sequences for learning intrinsic images
% , containing both indoor and
% outdoor scenes
(Sec.~\ref{sec:dataset}), and a new
% deep network
approach to learning single-view intrinsic images from this data
(Sec.~\ref{sec:approach}).  Because we train from image sequences,
one learning approach would be to use existing sequence-based
intrinsic images algorithms to produce approximate ground truth
decompositions,
% for the input sequences,
then use these algorithmic outputs as supervision. However, we found
that for many image sequences, existing sequence-based algorithms
perform poorly because their assumptions are not met, as discussed in
Sec.~\ref{sec:dataset}. Hence, during training, our CNN directly takes
an image sequence as input, and processes it in a feed-forward fashion
to produce reflectance and shading for each image in the sequence, as
shown in Figure~\ref{fig:network}. Because the network processes each
image independently, at test time multiple images are not required,
i.e., we can use the network to produce a decomposition for a single
image.
%%%%
During training, the input images interact through our novel loss
function (Sec.~\ref{sec:approach}), which evaluates the predicted
decompositions jointly for the entire sequence. 
%Our loss function
%includes terms measuring image reconstruction error, consistency of
%the predicted reflectances across the sequence, and smoothness of
%reflectance and shading outputs.

% \zhengqi{do we need to say our loss is sophisticated? Compared with
%  Jon Barron's paper, our energy is way simpler. I changed a little
%  bit}
%, and includes terms measuring image reconstruction
%error, the consistency of the reflectance images produced for the
%sequence, and smoothness for both reflectance and shading outputs.

For our network, we use a variant of the U-Net
architecture~\cite{ronneberger2015u, isola2016image}
(Figure~\ref{fig:network}). Our network has one encoder and two
decoders, one for log-reflectance and one for log-shading, with skip
connections for both decoders.
%so that the network can capture the
%details in the input images.
%%%%%%%%%%
Each layer of the encoder consists mainly of a $4\times 4$ stride-2
convolutional layer followed by batch
normalization~\cite{ioffe2015batch} as well as leaky
ReLu~\cite{he2015delving}. For the two decoders, each layer is
composed of a $4\times 4$ deconvolutional layer followed by ReLu.
%% In our network architecture, one decoder is used to predict log
%% reflectance and another one is used to predict grayscale log
%% shading.
In addition to the decoders for reflectance and shading, the network
predicts one side output from the innermost feature maps, a single RGB
vector for each image corresponding to the predicted illumination
color.
% in log-space.

%Our work has two main features: a large dataset of depth maps derived
%from multi-view stereo (Section~\ref{sec:dataset}), and an approach to
%learning to predict depth from this data (Section~\ref{sec:network}).
%\noah{This section is short. I think we can drop it, and just move
%  this text to the introduction if it isn't there already.}

\section{Dataset}\label{sec:dataset} 

To create the \BT dataset, we collected videos and image sequences
depicting both indoor and outdoor scenes with varying
illumination. While many time-lapse datasets primarily capture outdoor
scenes, we explicitly wanted representation from indoor scenes as well. Our
indoor sequences were gathered
from Youtube, Vimeo, Flickr, Shutterstock, and
Boyadzhiev~\etal~\cite{Boyadzhiev2013UserassistedIC},
and our outdoor sequences were collected
from the AMOS~\cite{jacobs2007consistent} and Time
Hallucination~\cite{shih2013data} datasets. For each video, we masked
out the sky as well as dynamic objects such as pets, people, and
cars via automatic semantic segmentation~\cite{zhao2016pyramid} or
manual annotation. We collected 145 sequences from indoor scenes and 50 from outdoor scenes, yielding a total of $\sim$6,500 training images.

\begin{figure}[t]
  \centering
    \begin{tabular}{@{\hspace{0.0em}}c@{\hspace{0.0em}}c@{\hspace{0.0em}}c@{\hspace{0.0em}}c@{\hspace{0.0em}}}
        \includegraphics[width=0.24\columnwidth]{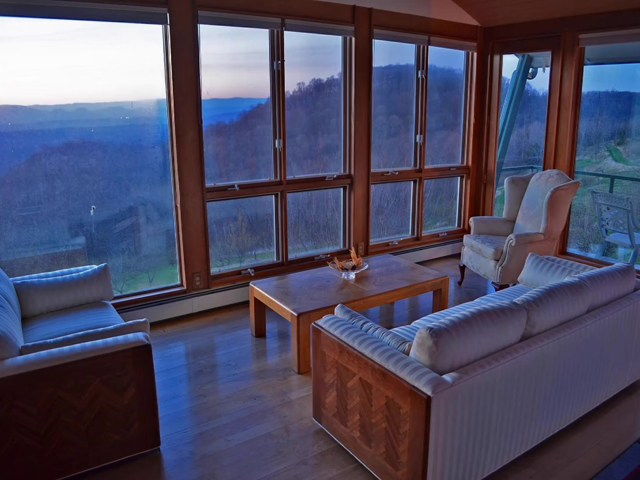} \vspace{-0.1em}   & 
        \includegraphics[width=0.24\columnwidth]{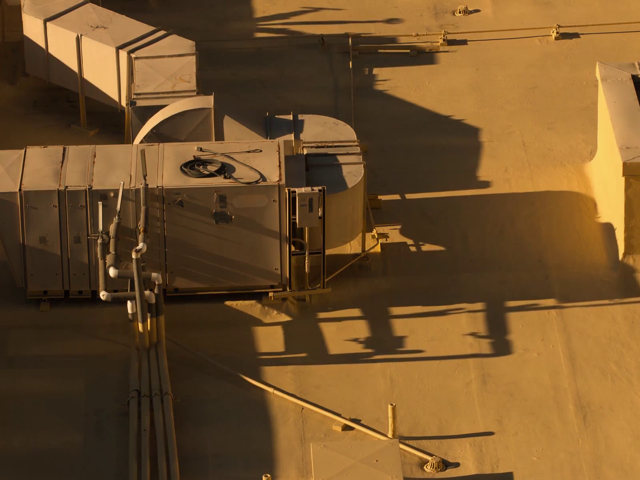}  \vspace{-0.1em}   &
        \includegraphics[width=0.24\columnwidth]{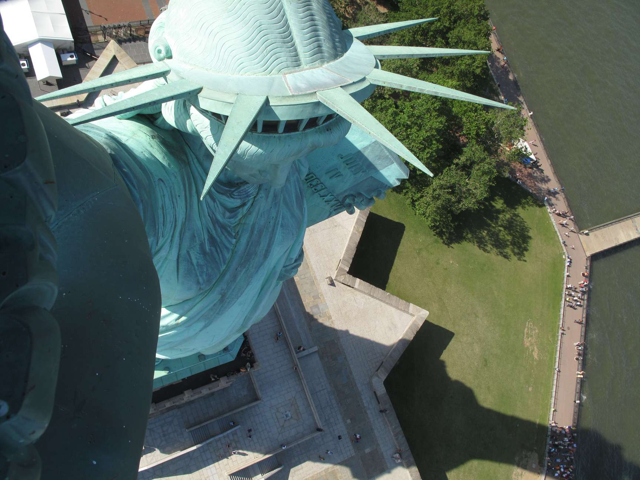} \vspace{-0.1em}   & 
        \includegraphics[width=0.24\columnwidth]{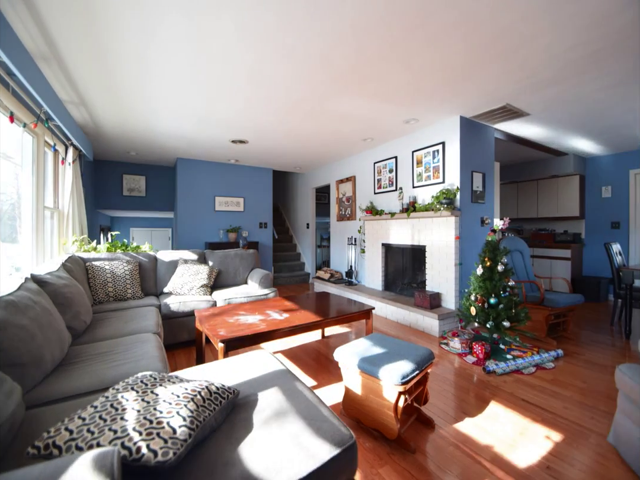} \vspace{-0.1em}  
    \end{tabular} 
  \caption{\textbf{Examples of challenging images in our dataset.} The
    first two images depict colorful illumination. The last two images
    show strong sunlight/shadows.}\label{fig:challenge_dataset}
    \vspace{-0.5em}
\end{figure}

\begin{figure}[t]
  \centering
    \begin{tabular}{@{\hspace{0.0em}}c@{\hspace{0.0em}}c@{\hspace{0.0em}}c@{\hspace{0.0em}}}
        \includegraphics[width=0.3\columnwidth]{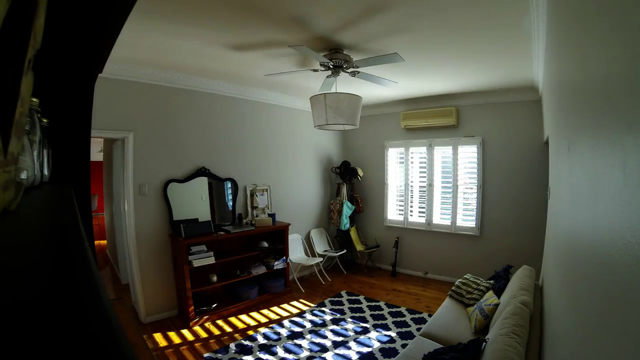} \vspace{0.0em} & 
        \includegraphics[width=0.3\columnwidth]{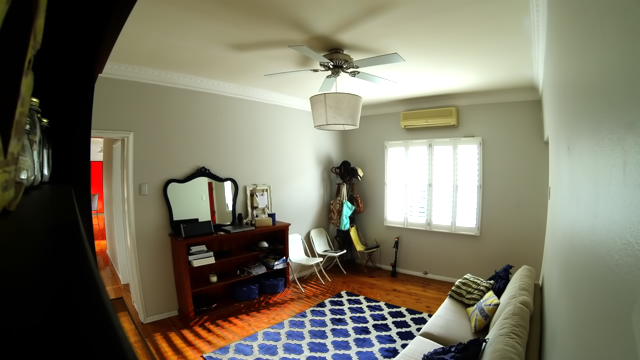}  \vspace{0.0em} &
        \includegraphics[width=0.3\columnwidth]{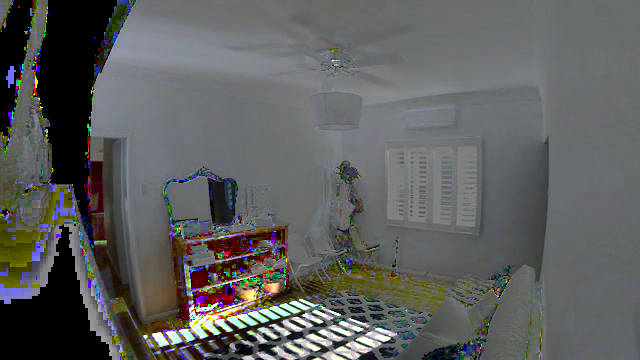} \vspace{0.0em}\\   
        \includegraphics[width=0.3\columnwidth]{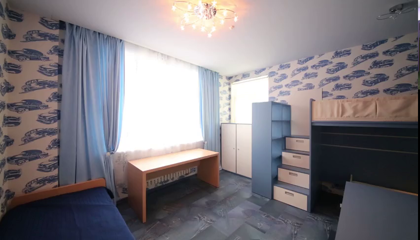} \vspace{-0.1em} &
        \includegraphics[width=0.3\columnwidth]{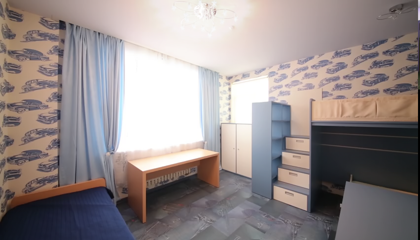} \vspace{-0.1em}&
        \includegraphics[width=0.3\columnwidth]{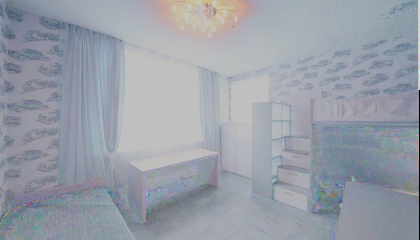} \vspace{-0.1em}\\
        {\scriptsize Image} & {\scriptsize Estimated $R$} & {\scriptsize Estimated $S$} \vspace{-0.5em}
    \end{tabular} 
  \caption{ \textbf{Failure cases for intrinsic image estimation
      algorithms.} We applied a state-of-the-art multi-image intrinsic
    image decomposition estimation
    algorithm~\cite{hauagge2013photometric} to our dataset. This
    method fails to produce decomposition results suitable for
    training due to strong assumptions that hold primarily for
    outdoor/laboratory scenes.
    % \noah{Should we also try Weiss's
    %  algorithm?} \zhengqi{their code only work grayscale images}
    \label{fig:fail}}
\end{figure}

\smallskip
\noindent{\bf Challenges with Internet videos.} Most outdoor scenes in
our dataset are from time-lapse sequences where the sun moves evenly
over time. Many existing algorithms for multi-image intrinsic image
decomposition work well on such data.
However, we found that indoor image sequences
are much more challenging because illumination changes in indoor scenes tend to be less even or continuous compared to outdoor scenes. In particular, we observed
that:
%%%%%%
\begin{packed_enum}
\item most relevant video clips cover a short period of time and do
  not show large changes in light direction,
\item several video clips are comprised of a light turning on/off in a
  room, producing a limited number ($<$8) of valid images with
  different lighting conditions, and
\item the dynamic range of indoor scenes can be high, with strong
  sunlight or shadows leading to saturation/clipping that can break
  intrinsic image algorithms.
\end{packed_enum}
%%%%%
% In fact, 
These properties make our dataset even more complex
% and challenging
than the IIW and SAW datasets. Several difficult examples are shown in
Fig.~\ref{fig:challenge_dataset}. We found that prior intrinsic image
methods designed for image sequences often fail on our indoor videos,
as their assumptions tend to hold only for outdoor or lab-captured
sequences. Example failure cases are shown in
Fig.~\ref{fig:fail}. However, as we show in our evaluation, our
approach is robust to such strong illumination conditions, and
networks trained on \BTShort generalize well to IIW and SAW.

\section{Approach}\label{sec:approach}

In this section, we describe our novel framework for learning
reflectance and shading from Internet time-lapse video clips. During
training, we formulate the problem as a continuous densely connected
conditional random field (dense CRF) and learn a deep neural network
to directly predict a decomposition from single views in a
feed-forward fashion.

\smallskip
\noindent{\bf Image formation model.} Let $I$ denote an input image,
and $R$ and $S$ denote the predicted reflectance (albedo) and
shading. Assuming an image of a Lambertian scene, we can write the
image decomposition in the log domain as:
\begin{align}
  \log I = \log R + \log S + N
\label{eq:formation}
\end{align}
where $N$ models image noise as well as deviations from a Lambertian
assumption. In our model, $S$ is a single-channel (grayscale) image,
while $R$ is an RGB image.
However, modeling $S$ with a single channel assumes white light. In
practice, the illumination color can vary across each input video (for
instance, red illumination at sunset/sunrise).
% , and such coloring is not always corrected to white by the camera.
Hence, we also allow for a colored light in our model:
\begin{align}
  \log I = \log R + \log S + c + N
\label{eq:formation_with_color}
\end{align}
where $c$ is a single RGB vector that is added to each element of the
left-hand side.
For simplicity, we use Eq.~\ref{eq:formation} in the following
sections; without loss of generality, we treat $c$ as being folded
into the predicted shading.
Each training instance is a stack of $m$ input images with $n$ pixels taken from a fixed viewpoint and varying illumination. We denote such an image
sequence by $\Images = \left \{ I^i | i = 1 \ldots m \right \}$, and
denote the corresponding predicted reflectances and shadings by
$\Reflectances = \left \{ R^i | i = 1 \ldots m \right \}$, and
$\Shadings = \left \{ S^i | i = 1 \ldots m \right \}$,
respectively. Additionally, for each image $I^i$ we have a binary mask
$\Mask^i$ indicating which pixels are valid (which we use to exclude
saturated pixels, sky, dynamic objects, etc).

We wish to devise a method for learning single-view intrinsic image
decomposition that leverages having multiple views during training.
Hence, we propose to combine learning and estimation
by encoding our priors into the training loss function. Essentially,
we learn a feed-forward predictor for single-image intrinsic images,
trained on image sequences with a loss that incorporates these priors,
and in particular priors that operate at the {\em sequence} level.
This loss should also be differentiable and efficient to evaluate,
considerations which guide our design below.

\medskip
\noindent{\bf Energy/loss function.}

During training, we formulate the problem as a dense CRF over an image
sequence $\Images$, where our goal is to maximize a posterior
probability $p(\Reflectances, \Shadings | \Images) = \frac{1}{Z(\Images)}
  \exp\left( - \Energy(\Reflectances, \Shadings, \Images)\right)$, where $Z(\Images)$ is the partition function. Maximizing $p(\Reflectances, \Shadings | \Images)$ is equivalent to minimize an energy function $\Energy(\Reflectances,
\Shadings, \Images)$. Because we use a feed-forward network to predict
the decomposition, we also use this energy
function as our training loss. We define $\Energy$ as:
\begin{align}
  \Energy(\Reflectances, \Shadings, \Images) = & \Limrec + \wrc
  \Lrc + \wrsm \Lrsm  \nonumber \\ 
  & +  \wssm \Lssm  \label{eq:loss}
\end{align}

We now describe each term in Eq.~\ref{eq:loss} in detail.

\subsection{Image reconstruction loss}\label{sec:imrec}
Given an input sequence $\Images$, for each image $I^i \in \Images$ we
expect the predicted reflectance and shading for $I^i$ to
approximately reconstruct $I^i$ via our image formation
model. Moreover, since reflectance is constant over time, we should
be able to use the reflectance $R^j$ predicted for {\em any} image
$I^j\in \Images$ to reconstruct $I^i$, when paired with $S^i$ (and
masked by the valid image regions indicated by binary masks $M^i$ and
$M^j$). This yields a term involving all pairs of images:
\small
\begin{multline}
  \Limrec = \\
  \sum_{i=1}^{m} \sum_{j=1}^{m} \norm{L^i \otimes \Mask^i
    \otimes \Mask^j \otimes (\log I^i - \log R^j - \log S^i)}^2_F
\end{multline}
\normalsize
%%%%%%%%%%%

where $\otimes$ is the Hadamard
product. Similar to~\cite{chen2013simple}, we weight our
reconstruction loss by input pixel luminance $L^i =
\text{lum}(I^i)^{\frac{1}{8}}$, since dark pixels tend to be noisy,
and image differences in dark regions are magnified in log-space.

We found that including such an {\em all-pairs connected} image
reconstruction loss improves prediction results, perhaps because it
creates more communication between predictions. A direct
implementation of this loss takes time $O(m^2n)$. In
Sec.~\ref{sec:trick} we introduce a computational trick that reduces
this to $O(mn)$ time, which is key to making training tractable.

\subsection{Reflectance consistency}\label{sec:refcon}
We also include a {\em reflectance consistency} loss that directly
encodes the assumption that the predicted reflectances should be
identical across the image sequence:
%%%%%%
\begin{equation}
  \Lrc = 
  \sum_{i=1}^{m} \sum_{j=1}^{m} \norm{\Mask^i \otimes \Mask^j
    \otimes (\log R^i - \log R^j)}^2_F\label{eq:cons}
\end{equation}
As above, this can be directly computed in time $O(m^2n)$, but
Sec.~\ref{sec:trick} shows how to reduce this to $O(mn)$.

\subsection{Dense spatio-temporal reflectance smoothness}\label{sec:rsmooth}
Our reflectance smoothness term $\Lrsm$ is based on the similarity of
chromaticity and intensity between pixels. Because we see a {\em
  sequence} of images at training time, we can define a reflectance
smoothness term that acts {\em jointly} on all of the images in each
sequence at once, allowing us to express smoothness in a richer way.
Accordingly, we introduce a novel spatio-temporal densely connected
reflectance smoothness term that considers the similarity of the
predicted reflectance at each pixel in the sequence to \emph{all}
other pixels in the sequence. Our method is inspired by the
bilateral-space stereo method of Barron~\etal~\cite{barron2015fast},
but we show how to apply their single-image dense solver to an entire
image sequence and how to implement it inside a deep network. We
define our smoothness term as:

\begin{equation}
 \Lrsm = \frac{1}{2} \sum_{I^i, I^j} % \in \Images \times \Images}
 \sum_{\substack{p \in I^i\\q \in I^j}} \hat{W}_{pq}(\log R^i_p - \log
 R^j_q)^2 \label{e7}\\ 
\end{equation}
%%%%%%%
where $p$ and $q$ indicate pixels in the image sequence, and $\hat{W}$
is a (bistochastic) weight matrix capturing the affinity between any
two pixels $p$ and $q$. Computing this equation directly is very
expensive because it involves all pairs of pixels in the sequence, hence
we need a more efficient approach.

First, note that if $\hat{W}$ is a bistochastic matrix, we can rewrite
Eq.~\ref{e7} in the following simplified matrix form:
\begin{equation}
\Lrsm = \mathbf{r}^\top (I - \hat{W}) \mathbf{r} \label{eq:e8}
\end{equation}
where $\mathbf{r}$ is a stacked vector representation (of length $mn$)
of all of the predicted log-reflectance images in the sequence:
$\mathbf{r} = [\mathbf{r}^1\ \mathbf{r}^2 \cdots \ \mathbf{r}^m ]^\top$,
where $\mathbf{r}^i$ is a vector containing the values in $\log R^i$. 
However, now we have a potentially dense affinity matrix $\hat{W} \in \mathbb{R}^{mn\times mn}$.
But we can approximately evaluate this term much more efficiently if
the pixel-wise affinities are Gaussian, i.e.,
%%%%%%
\begin{align}
 W_{pq} = \exp\left(- (\mathbf{f}_p- \mathbf{f}_q)^\top \Sigma^{-1}
 (\mathbf{f}_p - \mathbf{f}_q) )\right) \label{eq:wpq}
\end{align}
%%%%%%
where $\fp$ and $\fq$ are feature vectors for pixels $p$ and $q$
respectively, and $\Sigma$ is a covariance matrix. We can
approximately minimize Eq.~\ref{eq:e8} in bilateral space by
factorizing the Gaussian affinity matrix $ W \approx S^\top \bar{B} S$,
where $\bar{B} = B_0 B_1 \cdots B_{d} + B_{d} B_{d-1} \cdots B_0$ is a
symmetric matrix constructed as a product of sparse matrices representing
blur operations
% on the lattice
in bilateral space, $d$ is the dimension of feature vector
$\fp$, and $S$ is a sparse splat/slicing matrix that transforms between image
space and bilateral space.
%%%%%
Finally, let $\hat{W} = N W N $ be a bistochastic
representation of $W$,
where $N$ is a diagonal matrix that bistochasticizes $W$ % to $\hat{W}$
\cite{knight2014symmetry}.
%%%%
%%%%%%
%%%%%%
This bilateral embedding allows us to write the loss in
Eq.~\ref{eq:e8} as:
\begin{equation}
  \Lrsm \approx \mathbf{r}^\top (I - N S^\top \bar{B} S N ) \mathbf{r}      \label{eq:e9}
\end{equation}
Note that $\Lrsm$ is differentiable and $N$ and $S$ are both sparse
matrices that can be computed efficiently. Our final form of $\Lrsm$
(Eq.~\ref{eq:e9}) can be computed in time $O((d+1)mn)$, rather than
$O(m^2n^2)$.

We define the feature vector used to compute the affinities in
Eq.~\ref{eq:wpq} as $\fp = [\ x_p, y_p, I_p , c_1, c_2\ ]^\top$, 
%\begin{equation}
%  \fp = [\ x_p, y_p, I_p , c_1, c_2\ ]^\top
%\end{equation}
where $(x_p, y_p)$ is the spatial position of pixel $p$ in the image,
$I_p$ is the intensity of $p$, and $c_1 = \frac{R}{R+G+B}$ and $c_2 =
\frac{G}{R+G+B}$ are the first two elements of the $L_1$ chromaticity
of $p$.

\subsection{Multi-scale shading smoothness}
In addition to a reflectance smoothness term, our loss also
incorporates a shading smoothness term, $\Lssm$. This term is
summed over each predicted shading image: $\Lssm = \sum_{i=1}^{m} \Lssmi(S^i)$,
where $\Lssmi(S^i)$ is defined as a weighted $L_2$ term over
neighboring pixels:
\begin{align}
  \Lssmi(S^i) = \sum_{p \in I^i} \sum_{q \in N(p)} v_{pq} \left( \log
  S^i_p - \log S^i_q\right)^2
\end{align}
where $N(p)$ denotes the 8-connected neighborhood around pixel $p$,
and $v_{pq}$ is a weight on each edge. 
Our insight is to leverage all of the input images to compute the weights for each individual image. We are inspired by Weiss~\cite{weiss2001deriving}, who derives a multi-image intrinsic images algorithm based on {\em median image derivatives} over the sequence. Essentially, we expect the
median image derivative over the input sequence (in the log domain) to
approximate the derivative of the reflectance image.
If we denote $J_{pq} = \log I_p - \log I_q$ (dropping the
image index $i$ for convenience), then this suggests a weight of the
form:
%%%%%%%%
\begin{equation}
\vmedian_{pq} = \exp\left( -\vmedianweight \left( J_{pq}
-\median\{J_{pq}\} \right)^2 \right) \label{eq:badss}
\end{equation}
where $\median\{J_{pq}\}$ is the median value of $J_{pq}$ over the image
sequence, and $\vmedianweight$ is a parameter defining the strength of
$\vmedian_{pq}$.
%% REMOVE THIS PART
This weight discourages shading smoothness where the gradient of a
particular image is very different from the median (as would happen,
e.g., for a shadow boundary).

\begin{figure}[t]
  \centering
    \begin{tabular}{@{\hspace{0.0em}}c@{\hspace{0.0em}}c@{\hspace{0.0em}}c@{\hspace{0.0em}}c@{\hspace{0.0em}}}
        \includegraphics[width=0.23\columnwidth]{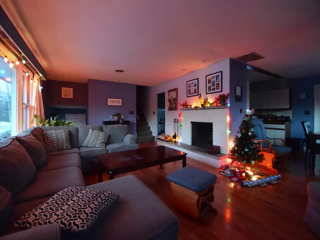} \vspace{-0.1em} &
        \includegraphics[width=0.23\columnwidth]{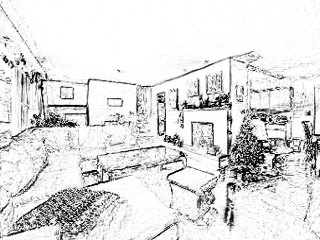} \vspace{-0.1em} &
        \includegraphics[width=0.23\columnwidth]{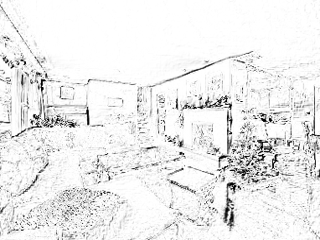} \vspace{-0.1em} &
        \includegraphics[width=0.23\columnwidth]{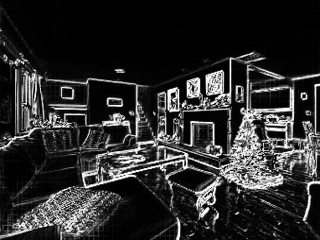} \vspace{-0.1em}
        \\ \includegraphics[width=0.23\columnwidth]{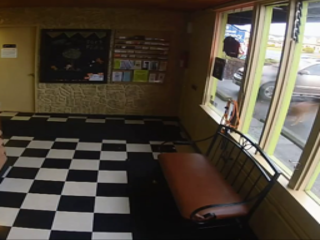} \vspace{-0.1em} &
        \includegraphics[width=0.23\columnwidth]{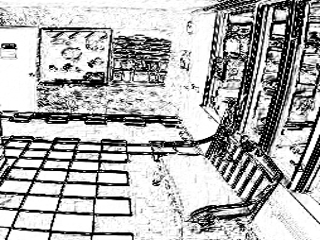} \vspace{-0.1em} &
        \includegraphics[width=0.23\columnwidth]{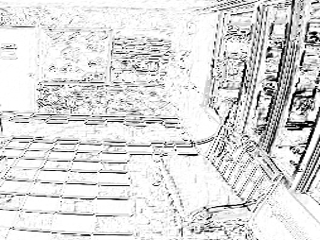} \vspace{-0.1em} 	&
        \includegraphics[width=0.23\columnwidth]{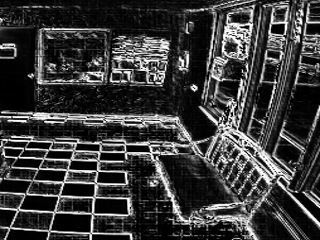} \vspace{-0.1em}
        \\ {\scriptsize Image} & {\scriptsize $\vmedian_{pq}$} &
           {\scriptsize $\max\{\vmedian_{pq}, \vmediannormpq\}$} &
           {\scriptsize $v_{pq}$ } \vspace{-0.5em}
    \end{tabular}  
  \caption{ \textbf{Effect of $\vmediannorm$ in shading smoothness
      term}. (white $=$ large weight, black $=$ small weight.) Adding
    the extra $\vmediannorm$ can help capture smoothness in textured
    regions such as the pillows in the first row and floor in the
    second row. The last column shows the final smoothness weight
    $v_{pq}$. \label{fig:msm}} \vspace{-1.0em}
\end{figure}
%%%%%%%
We found that $\vmedian_{pq}$ works well as a weight for texture-less
regions (for instance, it captures the effect of a cast shadow on a
flat wall well), but, due to noise present in dark image regions, it
does not always capture the desired shading smoothness for textured
surfaces. Figure~\ref{fig:msm} (bottom) illustrates such a case with a
checkerboard pattern on the floor. To address this issue, we define an
additional weight $\vmediannormpq$ that is normalized by the median
derivative:
\begin{equation}
  \vmediannormpq = \exp\left( -\vmediannormweight
  \left( \frac{J_{pq}-\median\{J_{pq}\}}{\median\{J_{pq}\}} \right)^2 \right) \label{eq:goodss}
\end{equation}
%%%%%%%%

We combine these weights as follows:
\begin{align}
  v_{pq} = \max\{\vmedian_{pq}, \vmediannormpq\} \cdot \left( 1 - \median\{W_{pq}\} \right)
\end{align}

This final shading smoothness weight is more robust to textured
regions while still distinguishing shadow discontinuities.  The last
factor $\left( 1 - \median\{W_{pq}\} \right)$ reflects the belief that
we should enforce stronger shading smoothness on reflectance edges
such as textures and weaker smoothness on regions of constant
reflectance. 

Ideally, our shading smoothness term would be densely
connected. However, the median operator is nonlinear and cannot be
integrated in a pixel-wise densely connected term. Instead, to
introduce longer-range shading constraints, we compute the shading
smoothness term at multiple image scales, by repeatedly downsizing
each predicted shading image by a factor of two.
We set the number of scales to be 4, and each scale $l$ is weighted by
a factor $\frac{1}{l}$.

\subsection{All-pairs weighted least squares (\acronym)}\label{sec:trick} 
Direct implementations of the all-pairs image reconstruction and
reflectance consistency terms from Sections~\ref{sec:imrec} and
\ref{sec:refcon} would take $O(m^2n)$ time.
This quadratic complexity would make training intractable for large
enough $m$.
Here, we propose a closed-form version of this all-pairs weighted
least squares loss (\acronym) that is linear in $m$.
While we apply this tool to our scenario, it can be used in other
situations involving all-pairs computation on image sequences.

In general, suppose each image $I^i$ is associated with two matrices
$P^i$ and $Q^i$ and two prediction images $X^i$ and $Y^i$. We then can
write \acronym as (see supplemental material for a detailed
derivation):

%%%%
\begin{align}
  \text{\acronym} = & \sum_{i=1}^m \sum_{j=1}^m  || P^i \otimes Q^j \otimes (X^i - Y^j)||^2_F \label{eq:slow_ls} \\ 
  = & \mathbf{1}^\top ( \Sigma_{Q^2} \otimes  \Sigma_{P^2X^2} + \Sigma_{P^2} \otimes  \Sigma_{Q^2Y^2} - \nonumber \\
  & 2 \Sigma_{P^2Y} \otimes \Sigma_{Q^2X} ) \mathbf{1} \label{eq:wdipls}
\end{align}
where
% the notation
$\Sigma_Z$ denotes the sum over all images of the
Hadamard product indicated in the subscript $Z$.
Evaluating Eq.~\ref{eq:slow_ls} requires time $O(m^2n)$, but rewritten
as Eq.~\ref{eq:wdipls}, just $O(mn)$.

We use this derivation to implement our image reconstruction loss
$\Limrec$ (Eq.~\ref{eq:wdipls}), by making the substitutions $P^i =
L^i \otimes M^i$, $Q^j = M^j$, $X^i = \log I^i - \log S^i$ and $Y^j =
\log R^j$, and our reflectance consistency loss $\Lrc$
(Eq.~\ref{eq:cons}) by substituting $P^i = M^i$, $Q^j = M^j$, $X^i =
\log R^i$ and $Y^j = \log R^j$.

\section{Evaluation}\label{sec:eval}
In this section we evaluate our approach by training solely on our \BT
dataset, and testing on two standard datasets, IIW and SAW. The
performance of machine learning approaches can suffer from
cross-dataset domain shift due to dataset bias. For example, we show
that the performance of networks trained on Sintel, MIT, or ShapeNet
do not generalize well to IIW and SAW.
However, our method, though \emph{not} trained on IIW or SAW data,
can still produce competitive results on both datasets.
We also evaluate on the MIT intrinsic images
dataset~\cite{grosse2009ground}, which has full ground truth. Rather
than using the ground truth during training, we train the network on
image sequences provided by the MIT dataset. 

\smallskip
\noindent{\bf Training details.} We implement our method in
PyTorch~\cite{pytorch}. In total, we have 195 image sequences for
training. We perform data augmentation via random rotations, flips,
and crops. When feeding images into the network, we resize
them to $256\times 384$, $384\times 256$, or $256\times 256$ depending
on the original aspect ratio. For all evaluations, we train the
network from scratch using Adam~\cite{Kingma2014AdamAM}.

\subsection{Evaluation on IIW}

%% IIW 
\begin{table}[t]
\centering
{\small
\begin{tabular}{llr}
  %% \hline
  \toprule
Method & Training set & WHDR\% \\
\midrule
%% \hline
Retinex-Color~\cite{grosse2009ground} & - & 26.9 \\
%% \hline 
Garces~\etal~\cite{garces2012intrinsic} & - & 24.8 \\
%% \hline 
%Shen and Yeo~\cite{shen2011intrinsic} & - & 32.5 \\
%% \hline 
Zhao~\etal~\cite{zhao2012closed} & - & 23.8 \\
%% \hline 
Bell~\etal~\cite{bell2014intrinsic} & - & 20.6  \\
\midrule
Narihira~\etal~\cite{narihira2015learning}${}^{*}$ & IIW & 18.1${}^{*}$  \\
Zhou~\etal~\cite{zhou2015learning}${}^{*}$ & IIW & 15.7${}^{*}$  \\
Zhou~\etal~\cite{zhou2015learning} & IIW & \textbf{19.9}  \\
\midrule
DI~\cite{narihira2015direct} & Sintel+MIT & 37.3 \\
\midrule
Shi~\etal~\cite{shi2016learning} & ShapeNet & 59.4 \\
\midrule
Ours (w/ per-image $\Limrec$) & \BTShort & 25.9 \\
%% \hline
Ours (w/ local $\Lrsm$) & \BTShort & 27.4 \\
%% \hline
Ours (w/ grayscale $S$) & \BTShort & 22.3 \\
%% \hline
Ours (full method)  & \BTShort & 20.3\\
\bottomrule
\end{tabular}
}
\caption{{\bf Results on the IIW test set.} Lower is better for the
  Weighted Human Disagreement Rate (WHDR). The second column indicates
  the training data each learning-based method uses; ``-'' indicates
   the method is optimization-based. ${}^{*}$ indicates WHDR is
   evaluated based on CNN classifer outputs for pairs of pixels rather
   than full decompositions.\label{tb:tb_IIW}} 
\vspace{-0.5em}
\end{table}

\begin{table}[t]
\centering
{\small
\begin{tabular}{llr}
\toprule
  Method & Training set & AP\% \\
\midrule
Retinex-Color~\cite{grosse2009ground} & - & 91.93 \\
%% \hline 
Garces~\etal~\cite{garces2012intrinsic} & - & 96.89 \\
%% \hline 
%Shen~\etal~\cite{shen2011intrinsic_optimization} & - & 94.12 \\
%% \hline 
Zhao~\etal~\cite{zhao2012closed} & - & 97.11 \\
%% \hline 
Bell~\etal~\cite{bell2014intrinsic} & - & 97.37  \\
\midrule
Zhou~\etal~\cite{zhou2015learning} & IIW & 96.24  \\
\midrule
DI~\cite{narihira2015direct} & Sintel+MIT & 95.04 \\
\midrule
Shi~\etal~\cite{shi2016learning} & ShapeNet & 86.30\\
\midrule
Ours (w/ local $\Lssm$)& \BTShort & {97.03} \\
Ours (w/o Eq.~\ref{eq:goodss}) & \BTShort & {97.15} \\
Ours (full method) & \BTShort & \textbf{97.90}\\
\bottomrule
\end{tabular}
}
\caption{{\bf Results on the SAW test set.}  Higher is better for
  AP\%. The second column is described in Table~\ref{tb:tb_IIW}. Note
  that none of the methods use annotations from SAW.  \label{tb:tb_SAW}}   \vspace{-0.5em}
\end{table}

\begin{figure*}[ptb]
 \centering
\centering
	\vspace*{0.1em} 
    \begin{subfigure}[b]{0.135\textwidth}
        \includegraphics[width=\textwidth]{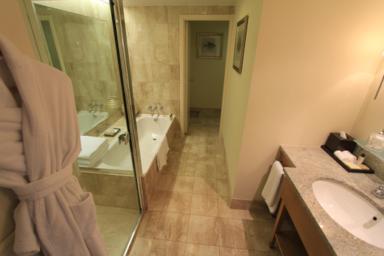}
   \end{subfigure} \hspace*{-0.8em}
    ~
    \begin{subfigure}[b]{0.135\textwidth}
        \includegraphics[width=\textwidth]{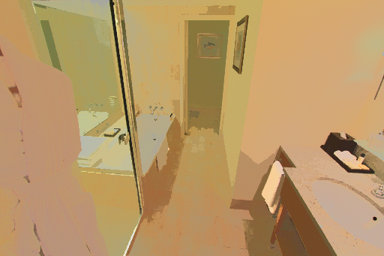}
    \end{subfigure} \hspace*{-0.8em}
    ~
    \begin{subfigure}[b]{0.135\textwidth}
       \includegraphics[width=\textwidth]{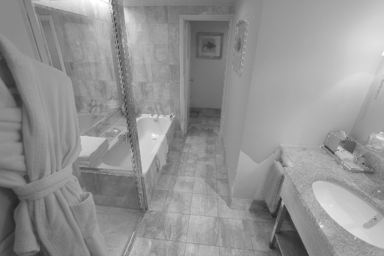}
    \end{subfigure} \hspace*{-0.8em}
    ~
    \begin{subfigure}[b]{0.135\textwidth}
        \includegraphics[width=\textwidth]{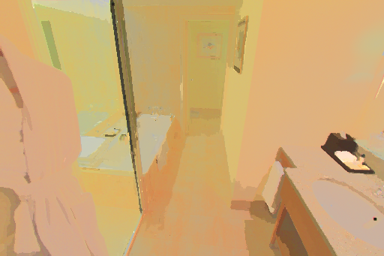}
    \end{subfigure} \hspace*{-0.8em}
    ~    
    \begin{subfigure}[b]{0.135\textwidth}
        \includegraphics[width=\textwidth]{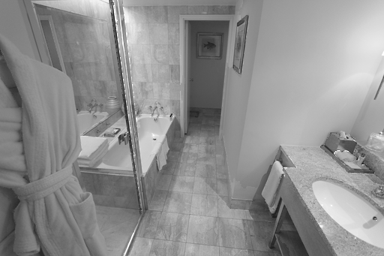}
    \end{subfigure}   \hspace*{-0.8em}
    ~
    \begin{subfigure}[b]{0.135\textwidth}
        \includegraphics[width=\textwidth]{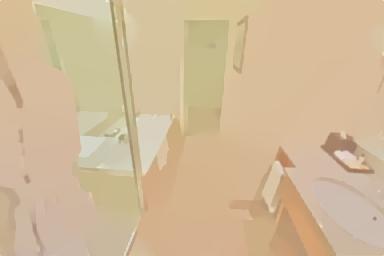}
       %\caption{Ours (seen)
       %\label{fig:corn_autopano}}
    \end{subfigure}   \hspace*{-0.8em}
    ~    
    \begin{subfigure}[b]{0.135\textwidth}
        \includegraphics[width=\textwidth]{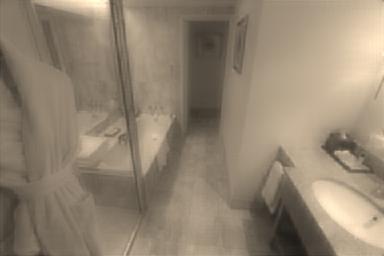}
       %\caption{Ours (unseen)
       %\label{fig:corn_autopano}}
    \end{subfigure} 
    %%%%%%%%%%%%%%%%%%%%%%%%%%%%%%%%%%%%%%%%%%%%%%%%%%%%%%%%%%%%%%%%%%%%%%%%%%%%%%%%%%
	% 2nd row 	 
	 \begin{subfigure}[b]{0.135\textwidth}
        \includegraphics[width=\textwidth]{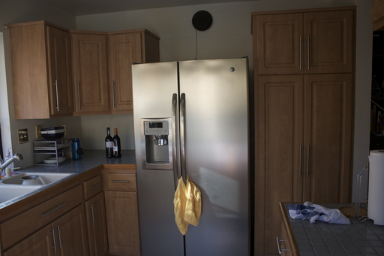}
        %\caption{Image
        %\label{fig:img1}}
    \end{subfigure} \hspace*{-0.8em}
    ~
    \begin{subfigure}[b]{0.135\textwidth}
        \includegraphics[width=\textwidth]{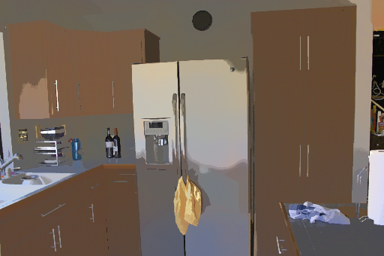}
        %\caption{ground truth
        %\label{fig:corn_autostitch}}
    \end{subfigure} \hspace*{-0.8em}
    ~
    \begin{subfigure}[b]{0.135\textwidth}
        \includegraphics[width=\textwidth]{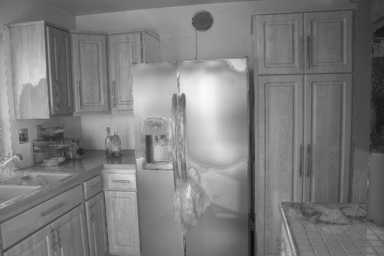}
        %\caption{DIW \cite{chen2016single}
        %\label{fig:corn_autopano}}
    \end{subfigure} \hspace*{-0.8em}
    ~
    \begin{subfigure}[b]{0.135\textwidth}
        \includegraphics[width=\textwidth]{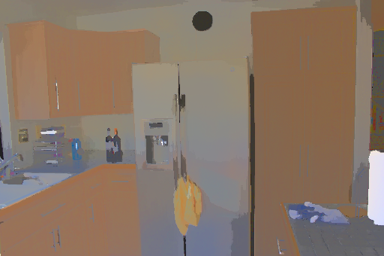}
        %\caption{DIW \cite{chen2016single}
        %\label{fig:corn_autopano}}
    \end{subfigure} \hspace*{-0.8em}
    ~    
    \begin{subfigure}[b]{0.135\textwidth}
        \includegraphics[width=\textwidth]{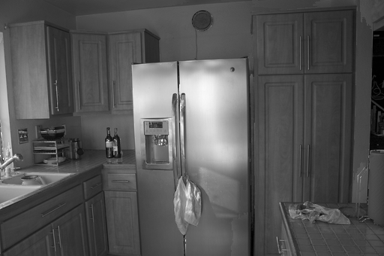}
        %\caption{Ours (seen)
        %\label{fig:corn_autopano}}
    \end{subfigure}   \hspace*{-0.8em}
    ~
    \begin{subfigure}[b]{0.135\textwidth}
        \includegraphics[width=\textwidth]{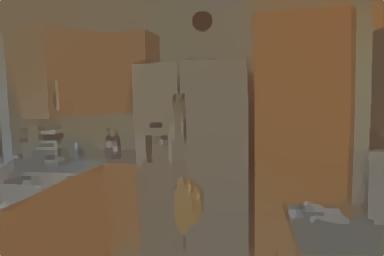}
        %\caption{Ours (seen)
        %\label{fig:corn_autopano}}
    \end{subfigure}   \hspace*{-0.8em}
    ~    
    \begin{subfigure}[b]{0.135\textwidth}
        \includegraphics[width=\textwidth]{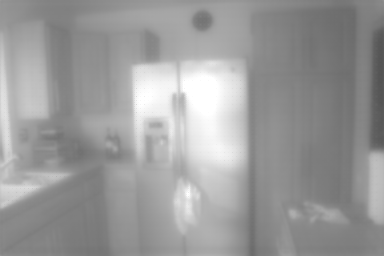}
        %\caption{Ours (unseen)
        %\label{fig:corn_autopano}}
    \end{subfigure} 
    %%%%%%%%%%%%%%%%%%%%%%%%%%%%%%%%%%%%%%%%%%%%%%%%%%%%%%%%%%%%%%%%%%%%%%%%%%%%%%%%%%
	% 3rd row 	 
	 \begin{subfigure}[b]{0.135\textwidth}
        \includegraphics[width=\textwidth]{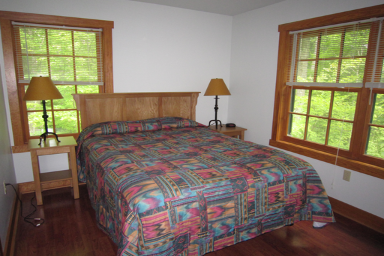}
        %\caption{Image
        %\label{fig:img1}}
    \end{subfigure} \hspace*{-0.8em}
    ~
    \begin{subfigure}[b]{0.135\textwidth}
        \includegraphics[width=\textwidth]{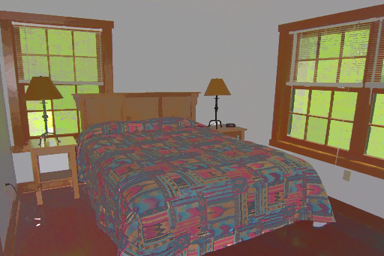}
        %\caption{ground truth
        %\label{fig:corn_autostitch}}
    \end{subfigure} \hspace*{-0.8em}
    ~
    \begin{subfigure}[b]{0.135\textwidth}
        \includegraphics[width=\textwidth]{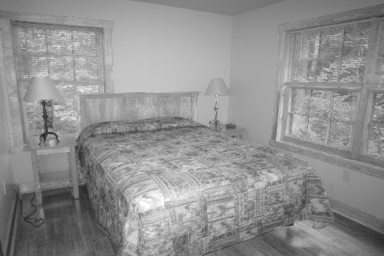}
        %\caption{DIW \cite{chen2016single}
        %\label{fig:corn_autopano}}
    \end{subfigure} \hspace*{-0.8em}
    ~
    \begin{subfigure}[b]{0.135\textwidth}
        \includegraphics[width=\textwidth]{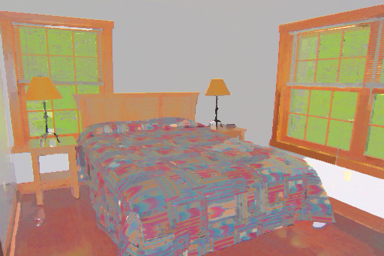}
        %\caption{DIW \cite{chen2016single}
        %\label{fig:corn_autopano}}
    \end{subfigure} \hspace*{-0.8em}
    ~    
    \begin{subfigure}[b]{0.135\textwidth}
        \includegraphics[width=\textwidth]{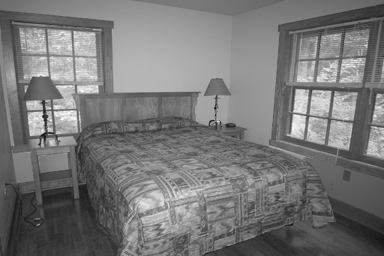}
        %\caption{Ours (seen)
        %\label{fig:corn_autopano}}
    \end{subfigure}   \hspace*{-0.8em}
    ~
    \begin{subfigure}[b]{0.135\textwidth}
        \includegraphics[width=\textwidth]{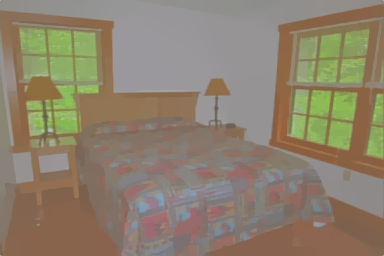}
        %\caption{Ours (seen)
        %\label{fig:corn_autopano}}
    \end{subfigure}   \hspace*{-0.8em}
    ~    
    \begin{subfigure}[b]{0.135\textwidth}
        \includegraphics[width=\textwidth]{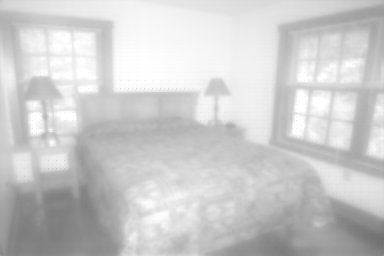}
        %\caption{Ours (unseen)
        %\label{fig:corn_autopano}}
    \end{subfigure}   
    %%%%%%%%%%%%%%%%%%%%%%%%%%%%%%%%%%%%%%%%%%%%%%%%%%%%%%%%%%%%%%%%%%%%%%%%%%%%%%%%%%
    %%%%%%%%%%%%%%%%%%%%%%%%%%%%%%%%%%%%%%%%%%%%%%%%%%%%%%%%%%%%%%%%%%%%%%%%%%%%%%%%%%
	% 8th row 	 
	 \begin{subfigure}[b]{0.135\textwidth}
        \includegraphics[width=\textwidth]{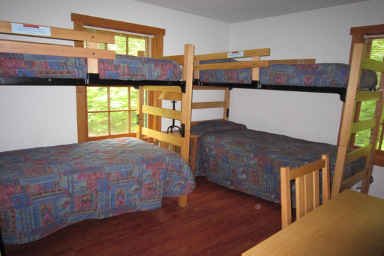}
        %\caption{Image
        %\label{fig:img1}}
         \caption{Image}  \vspace{-0.1em}
    \end{subfigure} \hspace*{-0.8em}
    ~
    \begin{subfigure}[b]{0.135\textwidth}
        \includegraphics[width=\textwidth]{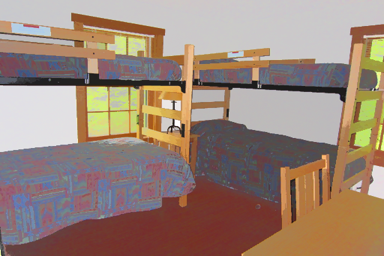}
        %\caption{ground truth
        %\label{fig:corn_autostitch}
         \caption{Bell~\etal (R)}  \vspace{-0.1em}
    \end{subfigure} \hspace*{-0.8em}
    ~
    \begin{subfigure}[b]{0.135\textwidth}
        \includegraphics[width=\textwidth]{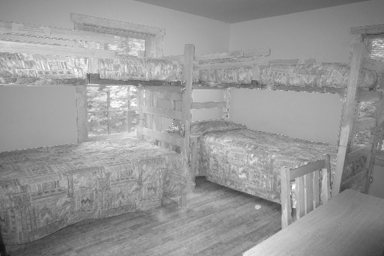}
        %\caption{DIW \cite{chen2016single}
        %\label{fig:corn_autopano}}
		\caption{Bell~\etal (S)}  \vspace{-0.1em}
    \end{subfigure} \hspace*{-0.8em}
    ~
    \begin{subfigure}[b]{0.135\textwidth}
        \includegraphics[width=\textwidth]{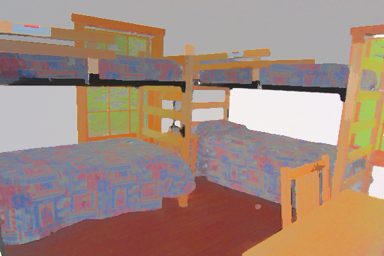}
        %\caption{DIW \cite{chen2016single}
        %\label{fig:corn_autopano}}
	\caption{Zhou~\etal (R)}  \vspace{-0.1em}
    \end{subfigure} \hspace*{-0.8em} 
    ~    
    \begin{subfigure}[b]{0.135\textwidth}
        \includegraphics[width=\textwidth]{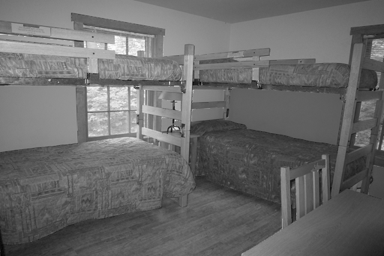}
        %\caption{Ours (seen)
        %\label{fig:corn_autopano}}
    \caption{Zhou~\etal (S)}  \vspace{-0.1em}
    \end{subfigure}   \hspace*{-0.8em}
    ~
    \begin{subfigure}[b]{0.135\textwidth}
        \includegraphics[width=\textwidth]{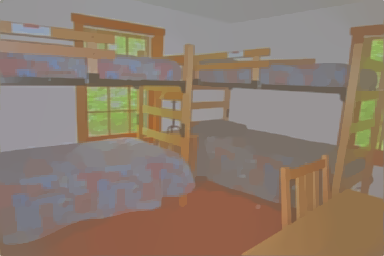}
        %\caption{Ours (seen)
        %\label{fig:corn_autopano}}
    \caption{Ours (R)}  \vspace{-0.1em}
    \end{subfigure}   \hspace*{-0.8em}
    ~    
    \begin{subfigure}[b]{0.135\textwidth}
        \includegraphics[width=\textwidth]{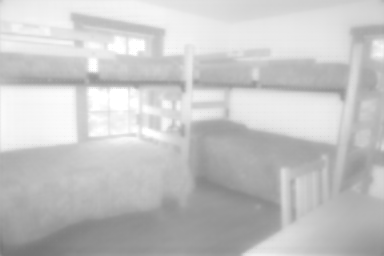}
    \caption{Ours (S)}  \vspace{-0.1em}
    \end{subfigure}  
    
  	\caption{\textbf{Qualitative comparisons for intrinsic image
            decomposition on the IIW/SAW test sets.} Our network
          predictions achieve comparable results to state-of-art
          intrinsic image decomposition algorithms (Bell~\etal
          ~\cite{bell2014intrinsic} and
          Zhou~\etal~\cite{zhou2015learning}).  \label{fig:visual_compare}}
        \vspace{-0.1em}%\vspace{-0.1em}
\end{figure*}

To evaluate on the IIW dataset, we train our network
on \BTShort (\emph{without} using IIW training data) and directly
apply our trained model on the IIW test split provided
by~\cite{narihira2015learning}. Numerical comparisons between our
method and other optimization-based and learning-based approaches are
shown in Table~\ref{tb:tb_IIW}.
Our method is competitive with both optimization-based
methods~\cite{bell2014intrinsic} and learning-based
methods~\cite{zhou2015learning}. Note that the best WHDR (marked
${}^*$) in the table is achieved using CNN classier outputs on pairs
of pixels, rather than full image decompositions. In contrast, our
results are based on full decompositions. Additionally, as we show in
the next subsection, the best performing method
(Zhou~\etal~\cite{zhou2015learning}) on IIW (which primarily evaluates
reflectance) falls behind on SAW (which evaluates shading), suggesting
that their method tends to overfit on reflectance. % , sacrificing
shading accuracy.
We also see that networks trained on Sintel, MIT or ShapeNet perform
poorly on IIW, likely due to dataset bias.

We also perform an ablation study on different configurations of our
framework. First, we modify the image reconstruction loss to an
alternate loss that considers each image independently, rather than
considering all pairs of images in a sequence. Second, we evaluate a
modified reflectance smoothness loss that uses local pairwise
smoothness (between neighboring pixels) rather than our proposed dense
spatio-temporal smoothness.
Finally, we try using grayscale shading, rather than our colored
shading. The results, shown in the last four rows of
Table~\ref{tb:tb_IIW}, demonstrate that our
full method can significantly improve reflectance predictions on the
IIW test set compared to simpler configurations.

\subsection{Evaluation on SAW}

\begin{table*}[tb]
\centering
{\small
\begin{tabular}{lllccccccccccc}
  %% \hline
  \toprule
&  & & \multicolumn{3}{c}{MSE} & \phantom{} & \multicolumn{3}{c}{LMSE}
& \phantom{} & \multicolumn{3}{c}{DSSIM} \\ 
\cmidrule{4-6} \cmidrule{8-10} \cmidrule{12-14}
Method & Training set & GT & refl. & shading & avg. & & refl. & shading & avg. & & refl. & shading & avg. \\
%% \hline
\midrule
SIRFS~\cite{barron2015shape} & MIT & Yes & \textbf{0.0147} &
\textbf{0.0083} & \textbf{0.0115} & & 0.0416 & \textbf{0.0168} &
\textbf{0.0292} & & \textbf{0.1238} & \textbf{0.0985} & \textbf{0.1111} \\
% \hline 
DI~\cite{narihira2015direct} & MIT+ST & Yes & 0.0277 & 0.0154 & 0.0215
& & 0.0585 & 0.0295 & 0.0440 & & 0.1526 & 0.1328 & 0.1427\\
% \hline
Shi~\cite{shi2016learning} & MIT+SN & Yes & 0.0278 & 0.0126 &
0.0202 & & 0.0503 & 0.0240 & 0.0372 & & 0.1465 & 0.1200 & 0.1332 \\
% \hline
Ours & MIT & No & \textbf{0.0147} & 0.0135 & 0.0141 & & \textbf{0.0341}
& 0.0253 & 0.0297 & & 0.1398 & 0.1266 & 0.1332\\
\bottomrule
\end{tabular} \vspace{-0.5em}
\caption{{\bf Results on MIT intrinsics. } For all error metrics,
  lower is better. ST$=$Sintel dataset and SN$=$ShapeNet dataset. The
  second column shows the dataset used for training. GT indicates
  whether the method uses ground truth for
  training. \label{tb:tb_MIT}} \vspace{-0.5em}
}
\end{table*}

Next, we test our network on SAW~\cite{kovacs2017shading}, again training \emph{without} using
data from SAW. We also propose two improvements to the metric used to
evaluate results on SAW:

First, the original SAW error metric
is based on classifying a pixel $p$ as having smooth/nonsmooth shading
based on the gradient magnitude of the predicted shading image, $||
\nabla S ||_2$, normalized to the range $[0,1]$. Instead, we measure
the gradient magnitude in the {\em log} domain. We do this because of
the scale ambiguity inherent to shading and reflectance, and because
it is possible to have very bright values in the shading channel
(e.g., due to strong sunlight), and in such cases if we normalize
shading to $[0,1]$ then most of the resulting values will be close to
0. In contrast, computing the gradient magnitude of log shading $||
\nabla \log S ||_2$ achieves scale invariance, resulting in fairer
comparisons for all methods. As in~\cite{kovacs2017shading}, we sweep a
threshold $\tau$ to create a precision-recall (PR) curve that captures
how well each method captures smooth and non-smooth shading.

Second, Kovacs~\etal~ \cite{kovacs2017shading} apply a $10\times 10$
maximum filter to the shading gradient magnitude image before
computing PR curves, because many shadow boundary annotations are not
precisely localized.
However, this maximum filter can result in degraded performance for
smooth shading regions. 
Instead, we use the max-filtered log-gradient-magnitude image when
classifying non-smooth annotations, but use the unfiltered log
gradient image when classifying smooth annotations (see supplementary
for details).

All methods, including our own, are trained without use of SAW data.
Average precision (AP) scores  are shown in Table~\ref{tb:tb_SAW} (please see the supplementary for
full precision-recall curves).
Our method has the best performance among all prior methods we tested,
and our full loss outperforms variants with terms removed. In
particular, our method outperforms the best optimization-based
algorithm~\cite{bell2014intrinsic} on {\em both} IIW and SAW. On the
other hand, Zhou~\etal~\cite{zhou2015learning} tends to overfit to
IIW, as their performance on SAW ranks lower than several other
methods. Again, networks trained on Sintel, MIT, and ShapeNet data
perform poorly on SAW.

\subsection{Qualitative results on IIW and SAW}
Figure~\ref{fig:visual_compare} shows qualitative results from our
method and two other state-of-art intrinsic image decomposition
algorithms, Zhou~\etal~\cite{zhou2015learning} and
Bell~\etal~\cite{bell2014intrinsic}, on test images from IIW and SAW.
Our results are visually comparable to these methods. One observation
is that our shading predictions for dark pixels can be quite dark,
leading to reduced contrast in the reflectance images. However, this loss of contrast does not hurt numerical performance. Additionally, like other CNN approaches~\cite{narihira2015direct, shi2016learning}, the direct predictions from our network may not strictly satisfy $I = R \cdot S$ since the two decoders predict $R$ and $S$ simultaneously at test time.  As future work, it would be interesting to use our predictions as priors for optimization to
address these issues.

\subsection{Evaluation on MIT intrinsic images}
The MIT intrinsic images dataset~\cite{grosse2009ground} contains 20
objects with ground truth reflectance and shading, as well as an
associated image sequence taken from 11 different directional light
sources. We use the same training-test split as in
Barron~\etal~\cite{barron2015shape}, but instead of training our
network on the ground truth provided by the MIT dataset, we train only
on the provided image sequences using our learning approach.
In this case, we configure our network to produce grayscale shading
outputs, since the MIT dataset only contains grayscale shading ground
truth images.

We compare our approach to several supervised learning methods
including SIRFS~\cite{barron2015shape}, Direct Intrinsics
(DI)~\cite{narihira2015direct} and Shi~\etal~\cite{shi2016learning}.
These prior methods all train using ground truth reflectance and
shading images, and additionally DI~\cite{narihira2015direct} and
Shi~\etal~\cite{shi2016learning} pretrain
% their networks
on Sintel~\cite{Butler:ECCV:2012} and
ShapeNet~\cite{chang2015shapenet}, respectively. In contrast,
% for our results,
we train our network from scratch and only use the provided image
sequences
% provided by the MIT dataset
during training. We adopt the same metrics as~\cite{shi2016learning},
including mean square error (MSE), local mean square error (LMSE), and
structural dissimilarity index (DSSIM).

\begin{figure}[t]
  \centering
    \begin{tabular}{@{\hspace{-0.1em}}c@{\hspace{-0.1em}}c@{\hspace{-0.1em}}c@{\hspace{-0.1em}}c@{\hspace{-0.1em}}c@{\hspace{-0.1em}}c@{\hspace{-0.1em}}}
        \includegraphics[width=0.145\columnwidth]{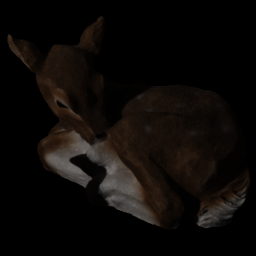} \vspace{-0.05em} & 
        \includegraphics[width=0.145\columnwidth]{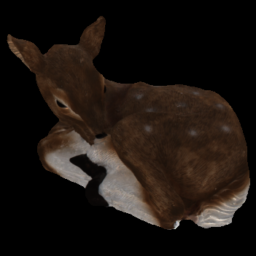}  \vspace{-0.05em} &
        \includegraphics[width=0.145\columnwidth]{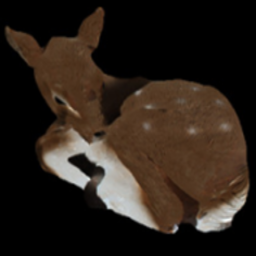}  \vspace{-0.05em} &
        \includegraphics[width=0.145\columnwidth]{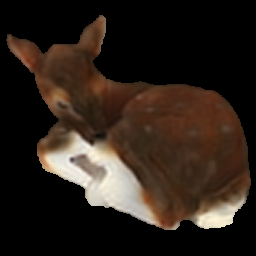}  \vspace{-0.05em} &
        \includegraphics[width=0.145\columnwidth]{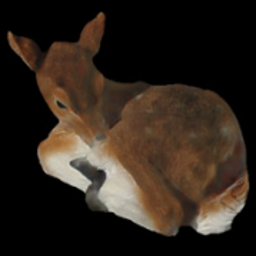}  \vspace{-0.05em} &
        \includegraphics[width=0.145\columnwidth]{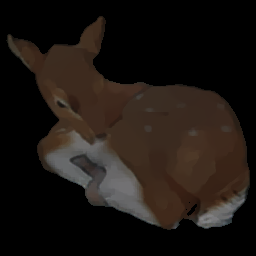}  \vspace{-0.05em} \\  
       \includegraphics[width=0.145\columnwidth]{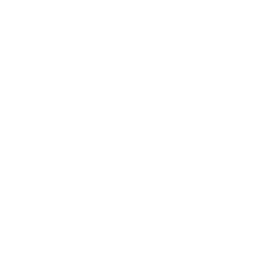} \vspace{-0.05em} & 
        \includegraphics[width=0.145\columnwidth]{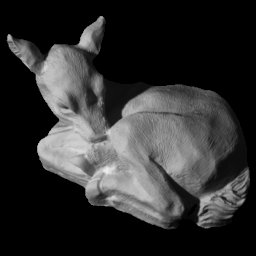}  \vspace{-0.05em} &
        \includegraphics[width=0.145\columnwidth]{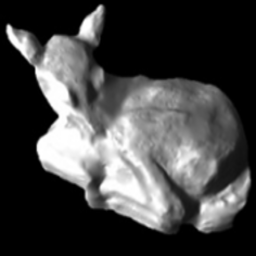}  \vspace{-0.05em} &
        \includegraphics[width=0.145\columnwidth]{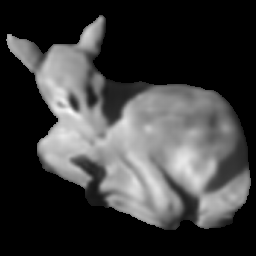}  \vspace{-0.05em} &
        \includegraphics[width=0.145\columnwidth]{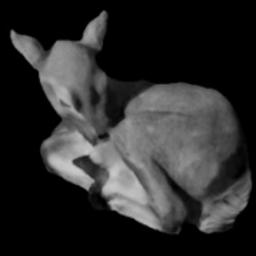}  \vspace{-0.05em} &
        \includegraphics[width=0.145\columnwidth]{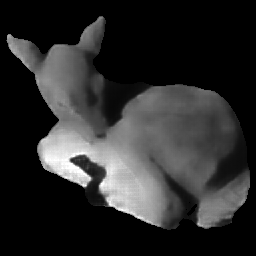}  \vspace{-0.05em} \\   
%% second image 
        \includegraphics[width=0.145\columnwidth]{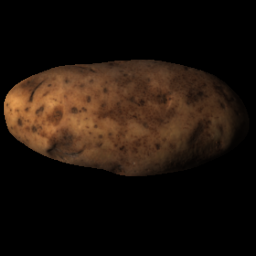} \vspace{-0.05em} & 
        \includegraphics[width=0.145\columnwidth]{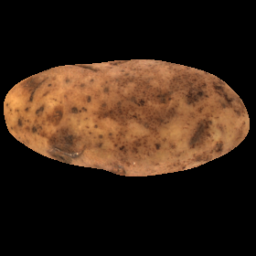}  \vspace{-0.05em} &
        \includegraphics[width=0.145\columnwidth]{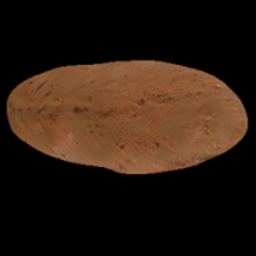}  \vspace{-0.05em} &
        \includegraphics[width=0.145\columnwidth]{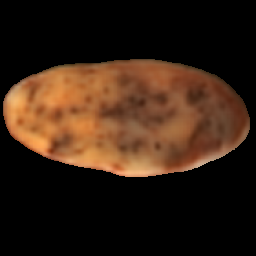}  \vspace{-0.05em} &
        \includegraphics[width=0.145\columnwidth]{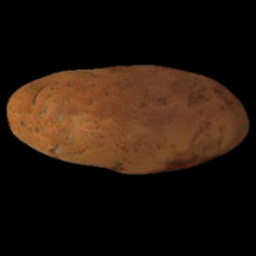}  \vspace{-0.05em} &        
        \includegraphics[width=0.145\columnwidth]{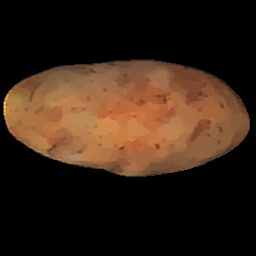}  \vspace{-0.05em} \\  
       \includegraphics[width=0.145\columnwidth]{figs/MIT/ones.png} \vspace{-0.05em} & 
        \includegraphics[width=0.145\columnwidth]{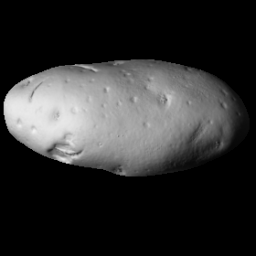}  \vspace{-0.05em} &
        \includegraphics[width=0.145\columnwidth]{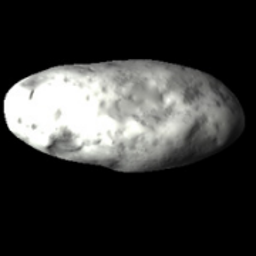}  \vspace{-0.05em} &
        \includegraphics[width=0.145\columnwidth]{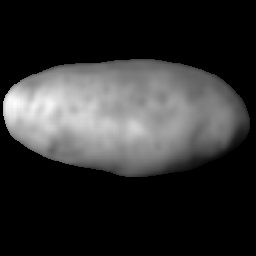}  \vspace{-0.05em} &
        \includegraphics[width=0.145\columnwidth]{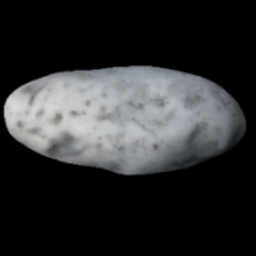}  \vspace{-0.05em} &
        \includegraphics[width=0.145\columnwidth]{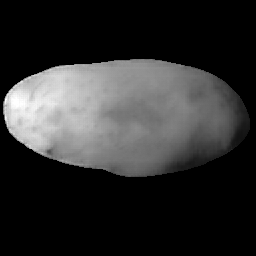}  \vspace{-0.05em} \\
%% third image
        \includegraphics[width=0.145\columnwidth]{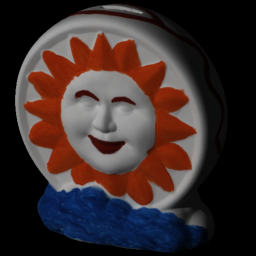} \vspace{-0.05em} & 
        \includegraphics[width=0.145\columnwidth]{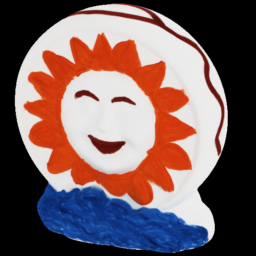}  \vspace{-0.05em} &
        \includegraphics[width=0.145\columnwidth]{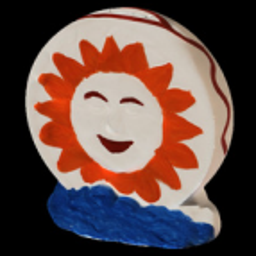}  \vspace{-0.05em} &
        \includegraphics[width=0.145\columnwidth]{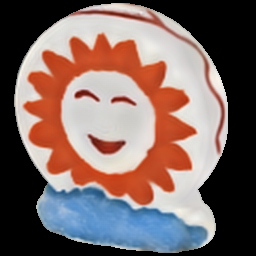}  \vspace{-0.05em} &
        \includegraphics[width=0.145\columnwidth]{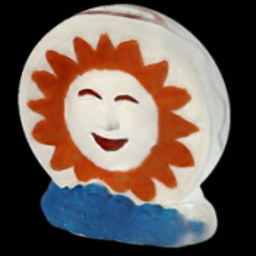}  \vspace{-0.05em} &         
        \includegraphics[width=0.145\columnwidth]{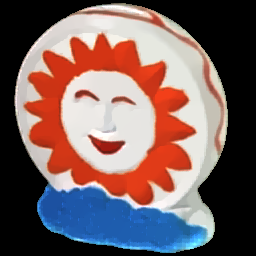}  \vspace{-0.05em} \\  
       \includegraphics[width=0.145\columnwidth]{figs/MIT/ones.png} \vspace{-0.05em} & 
        \includegraphics[width=0.145\columnwidth]{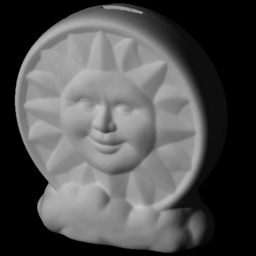}  \vspace{-0.05em} &
        \includegraphics[width=0.145\columnwidth]{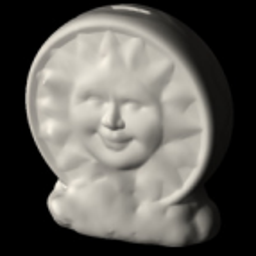}  \vspace{-0.05em} &
        \includegraphics[width=0.145\columnwidth]{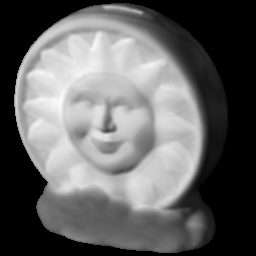}  \vspace{-0.05em} &
        \includegraphics[width=0.145\columnwidth]{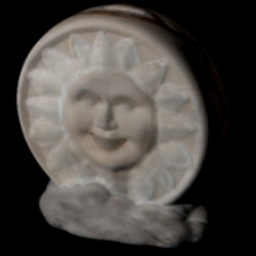}  \vspace{-0.05em}   &     
        \includegraphics[width=0.145\columnwidth]{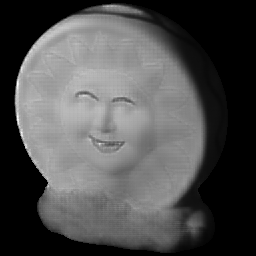}  \vspace{-0.05em} \\   
%% 4th image
%% third image
        \includegraphics[width=0.145\columnwidth]{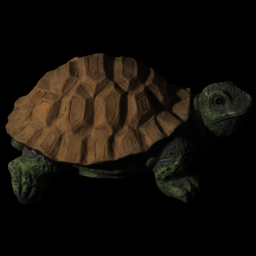} \vspace{-0.05em} & 
        \includegraphics[width=0.145\columnwidth]{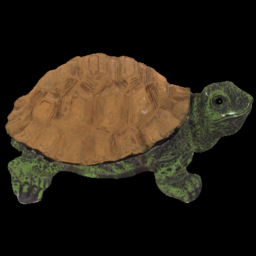}  \vspace{-0.05em} &
        \includegraphics[width=0.145\columnwidth]{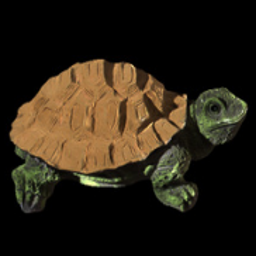}  \vspace{-0.05em} &
        \includegraphics[width=0.145\columnwidth]{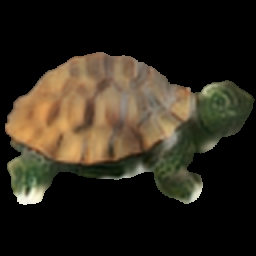}  \vspace{-0.05em} &
        \includegraphics[width=0.145\columnwidth]{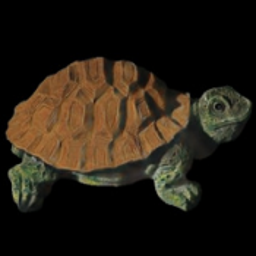}  \vspace{-0.05em}   &             
        \includegraphics[width=0.145\columnwidth]{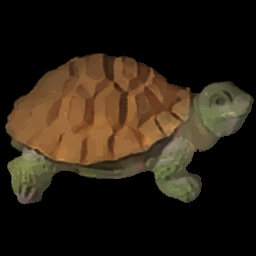}  \vspace{-0.05em} \\  
       \includegraphics[width=0.145\columnwidth]{figs/MIT/ones.png} \vspace{-0.05em} & 
        \includegraphics[width=0.145\columnwidth]{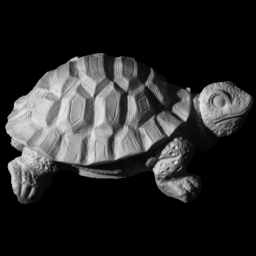}  \vspace{-0.05em} &
        \includegraphics[width=0.145\columnwidth]{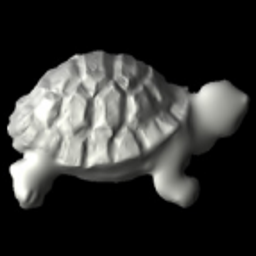}  \vspace{-0.05em} &
        \includegraphics[width=0.145\columnwidth]{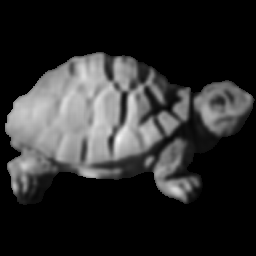}  \vspace{-0.05em} &
        \includegraphics[width=0.145\columnwidth]{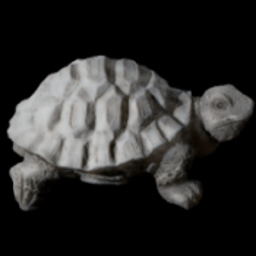}  \vspace{-0.05em}   &                     
        \includegraphics[width=0.145\columnwidth]{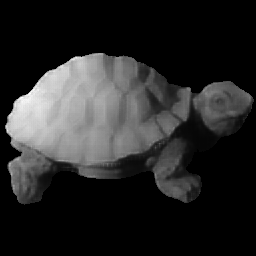}  \vspace{-0.05em} \\           
        {\scriptsize (a) Image} & {\scriptsize (b) GT} & {\scriptsize (c) SIRFS} & {\scriptsize (d) DI} &  {\scriptsize (e)Shi~\etal}  &  {\scriptsize (f)Ours} \vspace{-0.5em}
    \end{tabular} \vspace{-0.25em}
  \caption{ \textbf{Qualitative comparisons on the MIT intrinsic test
      set.} Odd-number rows show predicted reflectance; even-numbered
    rows show predicted shading. (a) Input image, (b) Ground truth
    (GT), (c) SIRFS~\cite{barron2015shape}, (d) Direct Intrinsics
    (DI)~\cite{narihira2015direct}, (e)
    Shi~\etal~\cite{shi2016learning}, (f) Our
    method. \label{fig:MIT}} \vspace{-0.5em}
\end{figure}

Numerical results are shown in Table~\ref{tb:tb_MIT} and qualitative
comparisons are shown in Figure~\ref{fig:MIT}.
%One can see that Our predictions
Averaged over reflectance and shading, our results numerically
outperform both prior CNN-based supervised learning
methods~\cite{narihira2015direct,shi2016learning}. In particular,
our albedo estimates are significantly better, while our shading
estimates are comparable (slightly better
than~\cite{narihira2015direct}, and slightly worse
than~\cite{shi2016learning}). SIRFS has the best numerical results on
the MIT, but SIRFS's priors only apply to single objects,
and their algorithm performs much more poorly on full images of
real-world scenes~\cite{narihira2015direct,shi2016learning}.

\section{Conclusion}
We presented a new method for learning intrinsic images, supervised
not by ground truth decompositions, but instead by simply observing
image sequences with varying illumination over time, and learning to
produce decompositions that are consistent with these sequences. Our
model can then be run on single images, producing competitive results
on several benchmarks. Our results illustrate the power of learning
decompositions simply from watching large amounts of video. In the
future, we plan to combine our approach with other kinds of
annotations (IIW, SAW, etc), to measure how well they perform when
used together, and to use our outputs as inputs to optimization-based
methods.

\smallskip
{\small
\noindent \textbf{Acknowledgments.} We thank Jingguang Zhou for his help with
data collection. We also thank the anonymous reviewers
for their valuable comments. This work was funded by the National
Science Foundation through grant IIS-1149393, and by a grant from
Schmidt Sciences.
}

\newpage
{\small
\bibliographystyle{ieee}
\bibliography{paper}
}

\end{document}